\documentclass[times,twocolumn,final,authoryear]{article}
\usepackage{simpleConference}
\usepackage{times}
\usepackage{graphicx}
\usepackage{amssymb}
\usepackage{url,hyperref}
\usepackage{xcolor}

\usepackage{natbib}
\usepackage[ngerman, english]{babel}
\usepackage{subcaption}
\usepackage{graphicx}
\usepackage{multirow}
\usepackage{array}
\usepackage{enumitem}
\setitemize{noitemsep,topsep=0.5pt,parsep=0.5pt,partopsep=0.5pt}
\newcolumntype{P}[1]{>{\centering\arraybackslash}p{#1}}
\usepackage{booktabs}
\newcommand\myuparrow{\mathord{\uparrow}}

\usepackage{authblk}
\author[1]{Chengwei Wei}
\author[2]{Bin Wang}
\author[1]{C.-C. Jay Kuo}
\affil[1]{University of Southern California, Los Angeles, California, USA}
\affil[2]{National University of Singapore, Singapore}

{
    \makeatletter
    \renewcommand\AB@affilsepx{: \protect\Affilfont}
    \makeatother

    \makeatletter
    \renewcommand\AB@affilsepx{, \protect\Affilfont}
    \makeatother

    \affil[ ]{\texttt{chengwei@usc.edu}}
}

\begin{document}

\title{SynWMD: Syntax-aware Word Mover's Distance for Sentence Similarity Evaluation}

% \author{Chengwei Wei\\
% \\
% University of Southern California \\
% Los Angeles, California, USA \\
% \today
% \\
% \\
% chengwei@usc.edu  \\
% }

\maketitle
\thispagestyle{empty}

\begin{abstract}
Word Mover's Distance (WMD) computes the distance between words and
models text similarity with the moving cost between words in two text
sequences. Yet, it does not offer good performance in sentence
similarity evaluation since it does not incorporate word importance and
fails to take inherent contextual and structural information in a
sentence into account. An improved WMD method using the syntactic parse
tree, called Syntax-aware Word Mover's Distance (SynWMD), is proposed to
address these two shortcomings in this work.  First, a weighted graph is
built upon the word co-occurrence statistics extracted from the
syntactic parse trees of sentences. The importance of each word is
inferred from graph connectivities. Second, the local syntactic parsing
structure of words is considered in computing the distance between
words.  To demonstrate the effectiveness of the proposed SynWMD, we
conduct experiments on 6 textual semantic similarity (STS) datasets and
4 sentence classification datasets. Experimental results show that
SynWMD achieves state-of-the-art performance on STS tasks. It also
outperforms other WMD-based methods on sentence classification tasks. 
\end{abstract}

\section{Introduction}\label{sec:introduction}

Sentence similarity evaluation has a wide range of applications in
natural language processing, such as semantic similarity computation
\citep{oliva2011symss}, text generation evaluation
\citep{zhao2019moverscore,bert-score}, and information retrieval
\citep{aliguliyev2009new, wang2015faq}. Methods for sentence similarity
evaluation can be categorized into two main classes: 1)
sentence-embedding-based methods and 2) word-alignment-based methods.
The former finds vector representations of sentences and calculates
the similarity of two sentences by applying a distance measure such as the cosine or $l_2$ distance. The latter operates at the word level and
uses the alignment cost of corresponding words in two sentences as the
sentence similarity measure. 

As one of the word-alignment-based methods, Word Mover's Distance (WMD) \cite{kusner2015word} formulates text similarity evaluation as a minimum-cost
flow problem.  It finds the most efficient way to align the information
between text sequences through a flow network defined by word-level
similarities. By assigning flows to individual words, WMD
computes text dissimilarity as the minimum cost of moving words' flows from
one sentence to another based on pre-trained word embeddings.  WMD is
interpretable as text dissimilarity is calculated as the distance
between words in two text sequences. 

However, a naive WMD method does not perform well on sentence similarity
evaluation for several reasons. First, WMD assigns word flow based on
words' frequency in a sentence. This frequency-based word weighting
scheme is weak in capturing word importance when considering the
statistics of the whole corpus.  Second, the distance between words
solely depends on the embedding of isolated words without considering
the contextual and structural information of input sentences.  Since the
meaning of a sentence depends on individual words as well as their
interaction, simply considering the alignment between individual words
is deficient in evaluating sentence similarity.  In this work, we
propose an enhanced WMD method called Syntax-aware Word Mover's Distance
(SynWMD). It exploits the structural information of sentences to improve
the naive WMD for sentence similarity evaluation. 

\begin{figure}[t]
\centering
\includegraphics[width=1.0\columnwidth]{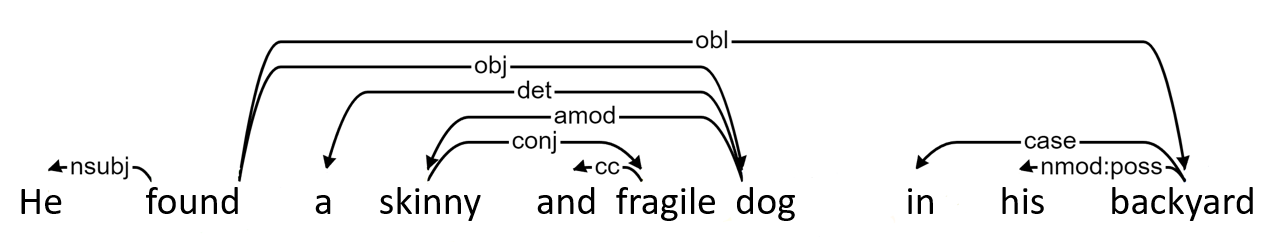}
\caption{The structure of a dependency parsing tree for an exemplary 
sentence: \emph{``He found a skinny and fragile dog in his backyard.''}}
\label{fig:parsing_tree}
\end{figure}

A syntactic parse tree represents a sentence using a tree structure. It
encodes the syntax information of words and the structure information of
a sentence. The dependency parse tree (see an example in Fig.
\ref{fig:parsing_tree}) is one type of the syntactic parse tree.  Each
node in the tree represents a word, and an edge represents the
dependency relation of two connected words.  Thus, words' related
contexts can be well captured by the structures of the dependency parse tree. For example, \emph{dog} in Fig. \ref{fig:parsing_tree} is one of the
most related contexts of \emph{found} as its objective.  Such a 
relationship can be easily spotted. In contrast, \emph{skinny} and
\emph{fragile} are not directly related to \emph{found} because they are
the modifiers of \emph{dog}. They are far away from \emph{found} in the
dependency parse tree, although they are close to \emph{found} in the
sequential order. The dependency parse tree provides valuable
information in semantic modeling and has been proven useful in various
NLP applications, such as word embedding \citep{levy2014dependency,
WEI2022}, semantic role labeling \citep{strubell2018linguistically},
machine translation \citep{nguyen2020tree}, and text similarity tasks
\cite{quan2019efficient, wang2020structural}. 

SynWMD incorporates the dependency parse tree technique in both word flow
assignment and word distance modeling to improve the performance on sentence similarity evaluation. This work has the following three major contributions.

%%%%%%%%%%%%%%%%%%%%%%%%%%%%%%%%%%%%%%%
%%%%%%%%%%%%%%%%%%%%%%%%%%%%%%%%%%%%%%%

\begin{itemize}
\item A new syntax-aware word flow calculation method is proposed.
Words are first represented as a weighted graph based on the
co-occurrence statistics obtained by dependency parsing trees. Then, a
PageRank-based algorithm is used to infer word importance. 
\item The word distance model in WMD is enhanced by the context
extracted from dependency parse trees. The contextual information of words and structural information of sentences are explicitly modeled as additional subtree embeddings. 
\item We conduct extensive experiments on semantic textual similarity
tasks and k-nearest neighbor sentence classification tasks to evaluate
the effectiveness of the proposed SynWMD. The code for SynWMD is
available at \url{https://github.com/amao0o0/SynWMD}. 
\end{itemize}
The rest of the paper is organized as follows. Related work is reviewed
in Sec. \ref{sec:review}. SynWMD is proposed in Sec. \ref{sec:method}.
Experimental results are shown in Sec. \ref{sec:experiments}. Finally,
concluding remarks are given in Sec. \ref{sec:conclusion}

\section{Related Work}\label{sec:review}

Recent studies on sentence similarity evaluation can be classified into
two main categories: sentence-embedding-based methods and
word-alignment-based methods. They are reviewed below. 

\subsection{Sentence Embedding}

One way to assess sentence similarity is through sentence embedding.
That is, a sentence is first encoded into a vector with an encoder.
The similarity of two sentences is then inferred from the distance of their
embedded vectors, where a simple distance metric such as the cosine or $l_2$ distance can be used. As to sentence embedding methods, a
simple and fast one is to pool word embeddings.  Several weighting
schemes \citep{arora2017simple, wang2020sbert} were adopted by
pooling-based methods for simple sentence embeddings. Yet, there is an
anisotropic problem in word-embedding-based pooling methods; namely, the
associated embeddings narrow in a cone region of the vector space
\citep{ethayarajh-2019-contextual}. It limits embeddings' capability in
sentence similarity evaluation.  To address this limitation,
post-processing techniques were proposed to alleviate the anisotropic
problem. For example, principal components removal \citep{mu2017all},
BERT-flow \citep{li2020sentence}, and BERT-whitening
\citep{su2021whitening} can make sentence embedding more uniformly
distributed so as to enhance the performance on sentence similarity
assessment. Recently, methods \citep{reimers2019sentence, gao2021simcse}
fine-tune pre-trained models on labeled data or use self-supervised
contrastive learning to achieve superior performance on sentence
similarity tasks. 

\subsection{Word Alignment}

Alignment-based methods measure the word matching degree for sentence
similarity evaluation. WMD is a popular alignment-based method. Its
extensions are widely used in text similarity tasks. For example,
Sentence Mover's Similarity targets the similarity measure of long and
multi-sentence text sequences \cite{clark2019sentence}. They use both
word embedding and sentence embedding to measure text similarity.  Word
Rotator's Distance \cite{yokoi2020word} shows that the norm of word
embedding encodes word importance while the angle between two word
embeddings captures word similarity. Consequently, they assign word flow
based on the norm of word embedding and compute the cosine distance for
the similarity measure.  Recursive Optimal Transport
\cite{wang2020structural} is a structure-aware WMD method. It uses a
binary or a dependency parse tree to partition a sentence into
substructures of multiple levels. Then, text similarity is
recursively calculated by applying WMD to substructures at the same
level. Yet, since there is no interaction between substructures at
different levels, its capability of sentence similarity measure can
be affected.

MoverScore \cite{zhao2019moverscore} and BERTScore \cite{bert-score} are
two newly developed alignment-based methods using contextual word
embeddings. Built upon the same concept as WMD, MoverScore uses the
Inverse Document Frequency (IDF) to assign word flow so that less
frequent words get higher flow weights. Furthermore, instead of adopting
static word embedding, it uses contextual word embedding. It
incorporates the word's contextual information in word embedding implicitly,
which enables the distance measure between words more accurately. Unlike
WMD which considers the matching degree between a word in one sentence
and all words in the other sentence, BERTScore uses the greedy match
between words, where each word is only matched to its most similar word
in the other sentence. Both MoverScore and BERTScore offer
state-of-the-art performance on text generation evaluation. 

\section{Proposed SynWMD}\label{sec:method}

We first briefly review WMD in this section. We then introduce two syntax-aware components in SynWMD to improve WMD. They are Syntax-aware Word Flow (SWF) and Syntax-aware Word Distance (SWD).

\subsection{Word Mover's Distance}\label{subsec:wmd}

Inspired by the Wasserstein metric, WMD measures text similarity using
the optimal transport distance. It first utilizes pre-trained word
embeddings to compute the distance between words in two text sequences.
Let $x_i$ be the embedding of word $i$. WMD defines the distance between
word $i$ and word $j$ as $c(i,j)=||x_i-x_j||_2$, which is also referred
to as the transport cost from word $i$ to word $j$. Next, WMD assigns a
flow to each word. The amount of flow $f_i$ of word $i$ is defined as the
normalized word occurrence rate in a single text sequence:
\begin{equation} \label{eq:wmd_wf}
f_i = \frac{\mbox{count}(i)}{|f|}, \quad |f|=\sum_i \mbox{count}(i).
\end{equation}
where $|f|$ is the total word count of a text sequence. Then, WMD
measures the dissimilarity of two texts as the minimum cumulative transport
cost of moving the flow of all words from one text sequence to the other. It
can be formulated as the following constrained optimization problem:
\begin{equation} \label{eq:wmd}
\min_{T_{ij} \ge 0} \sum_{i \in I}^{}\sum_{j \in J}^{} T_{ij}c(i,j)
\end{equation}
subject to:
\begin{equation} \label{eq:wmd_constraints}
\sum_{j \in J}^{} T_{ij}=f_i \; \forall i \in I, \mbox{  and  } 
\sum_{i \in I}^{} T_{ij}=f'_j  \;  \forall j \in J,
\end{equation}
where $I$ and $J$ are the sets of words in text sequence $d$ and $d'$,
respectively, and $T_{ij}$ represents the amount of the flow that travel
from word $i$ to word $j$, which is a variable to be determined.  The
above constrained optimization problem can be solved by linear
programming. 

WMD has two main shortcomings. First, important words in a sentence
should be assigned a higher flow in Eq. (\ref{eq:wmd}). Yet, WMD assigns
word flow according to word occurrence rate in a sentence. This simple
scheme cannot capture word importance in a sentence accurately. Second, the transport
cost between two words is solely decided by their word embeddings.
Nevertheless, the meaning of a word may be affected by its context and
the meaning of a sentence can be affected by the structure of word
combinations. It is desired to develop more effective schemes. They are
elaborated in Secs. \ref{subsec:wordflow} and \ref{subsec:worddistance}. 

\subsection{Syntax-aware Word Flow}\label{subsec:wordflow}

Important words in two sentences can largely decide their similarity.
As given in Eq. (\ref{eq:wmd}), a word with a higher flow has a greater
impact on WMD results. Thus, a more important word in a text sequence
should be assigned with a higher flow. We propose an enhanced word flow
assignment scheme, called syntax-aware word flow (SWD). Simply speaking,
we collect the co-occurrence times of words in dependency parse trees
from the whole dataset, and get word importance based on the
co-occurrence statistics for flow assignment. 

The computation of SWD is detailed below.
\begin{enumerate}
\item Parse all sentences in a dataset and count the co-occurrence time
of two words if they appear in a parse tree within $n$-hop.  The
co-occurrence count is further weighted by the distance between two
words in a parse tree; namely, it is divided by the hop number between
two words. 
\item Build a weighted graph for the dataset, where each node
corresponds to a word and the edge of two connected words has a weight of their co-occurrence time
as computed by Step 1. Clearly, words with higher edge weight frequently co-occur with other words in the dataset. They have less novelty and importance to the sentence. Based on this assumption, a word with a higher total edge weight should be assigned lower word flow.
% Clearly, words with more co-occurrence in the
% dependency parse tree have an edge of a higher weight. Edges work as
% function words that connect parts in sentences. Assign a word of a
% higher total edge weight a smaller flow. 
\item Use the weighted PageRank algorithm \citep{page1999pagerank} to
count all edge weights of a node, which gives a rough estimate of node
importance, and assign the inverse of the PageRank value as its word
flow. 
\end{enumerate}

The last step can be written mathematically as
\begin{eqnarray} \label{eq:swf}
PR(i) & = & (1-d)+d \cdot \sum_{j=1}^{} w_{ij} \frac{PR(j)}{\sum_{k=1}^{} w_{jk}}, \\
f_i   & = & \frac{1}{PR(i)},
\end{eqnarray}
where $w_{ij}$ is the edge weight between word $i$ and word $j$, $PR(i)$
represents the PageRank value of word $i$, and $d$ is a parameter used
to control the smoothness of word flow. In this way, SWF can assign a
word that co-occurs with other words in the parse tree more frequently a
lower flow. 

%%%%%%%%%%%%%%%%%%%%%%%%%%%%%%%%%%%%%%%
\begin{figure*}[t]
\centering
\includegraphics[width=0.9\linewidth]{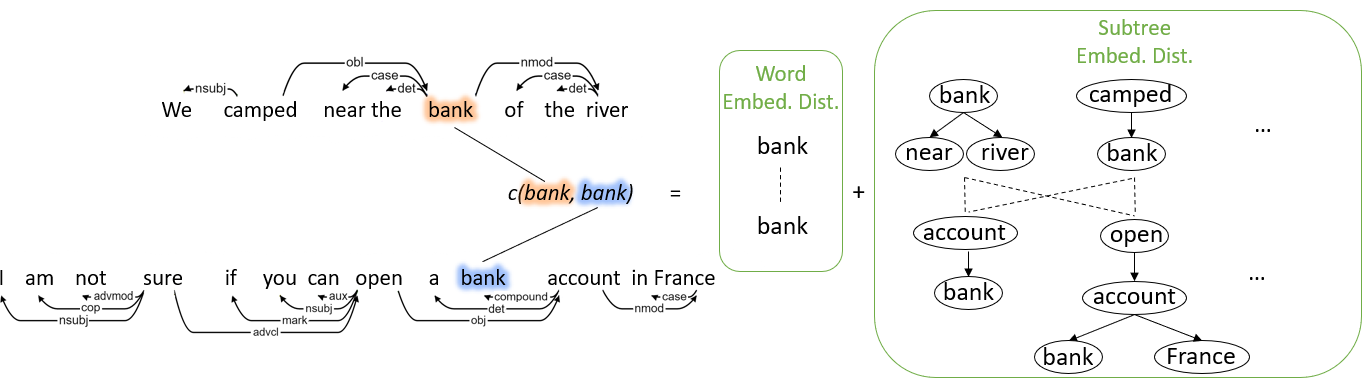}
\caption{Illustration of the shortcoming of distance calculation in WMD 
and an improved solution, SWD. In SWD, the distance between words is decided by both word embeddings and subtree embeddings}\label{fig:word_distance}
\end{figure*}
%%%%%%%%%%%%%%%%%%%%%%%%%%%%%%%%%%%%%%%

%%%%%%%%%%%%%%%%%%%%%%%%%%%%%%%%%%%%%%%
\begin{figure}[t]
\centering
\includegraphics[width=1\linewidth]{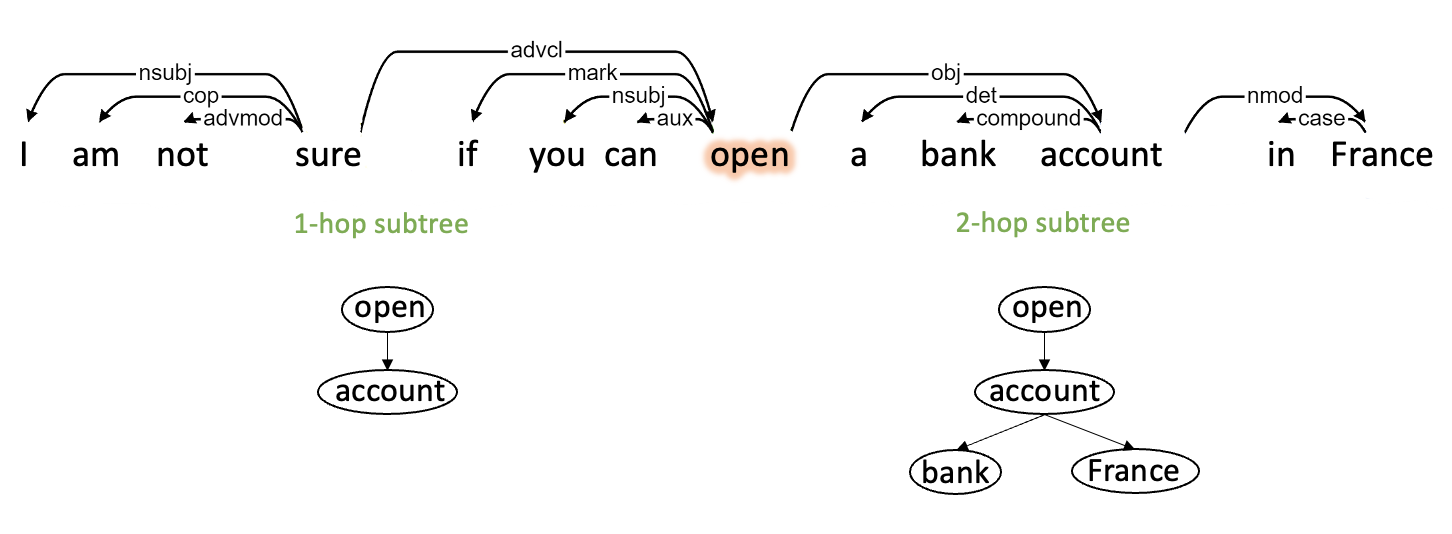}
\caption{1\&2-hop subtrees with \emph{open} as the parent node in the
dependency parsing tree for sentence: \emph{``I am not sure if you can
open a bank account in France''}. Stopwords are ignored.}\label{fig:subtree}
\end{figure}
%%%%%%%%%%%%%%%%%%%%%%%%%%%%%%%%%%%%%%%

\subsection{Syntax-aware Word Distance}\label{subsec:worddistance}

In WMD, the distance between words is called the transport cost. It is
computed using static word embedding without considering any word
contextual information or structural information of the sentence.  An
example given in Fig.  \ref{fig:word_distance} is used to illustrate
its shortcoming.  Two identical words \emph{bank} in the two sentences
have the distance of zero in WMD. However, they do not have the same
meaning because of different contexts. We can exploit contextual
word embedding to alleviate this problem, such as BERT-based models.
Here, we propose a syntax-aware word distance (SWD). Simply speaking,
SWD uses the dependency parse tree to find the most related context of
words and incorporates this information in word distance calculation. SWD
can be applied to both static and contextual word embeddings to improve
the performance of WMD. 

The procedure of SWD is detailed below.
\begin{enumerate}
\item Generate candidate subtrees from a dependency parse tree. For each
word in a tree, we treat it as a parent node, and use it and its
connections to \emph{m}-hop children to form the subtrees, where $m$ is
a hyper-parameter. With children from different hops, the context information from multiple levels can be extracted by the subtrees. Fig.  \ref{fig:subtree} shows 1-hop and 2-hop subtrees where the word "open" is the parent node. 
\item Collect all subtrees that contain the word as its context. Then
obtain the subtree embedding as the weighted average of all its word
embeddings. 
\item Incorporate the context of the target word in the word distance
calculation. As shown in Fig.  \ref{fig:word_distance}, besides distances between word embeddings, distances between subtree embeddings are also considered. 
\end{enumerate}

For the last step, the syntax-aware word distance between words $i$ and 
$j$ can be computed by
\begin{equation} \label{eq:swd}
c(i,j) = dist(v_i, v_j) + a \frac{\sum_{s_i\in S_i}^{}\sum_{s_j\in
S_j}^{} dist(s_i, s_j)}{|S_i|\cdot|S_j|},
\end{equation}
where $S_i$ and $S_j$ are the sets of subtrees that contain word $i$ and
$j$, respectively, and $a$ is a parameter controlling the amount of
contextual and structural information to be incorporated.  The cosine
distance,
\begin{equation}
dist(v_i, v_j)  = 1 - \frac{<v_i, v_j>}{||v_i|| \cdot ||v_j||},
\end{equation}
is used to measure the distance between word embeddings and subtree
embeddings.

\section{Experiments}\label{sec:experiments}

We evaluate SynWMD on six semantic textual similarity datasets and four
sentence classification datasets with the k-nearest neighbor classifier.
In all experiments, the sentence parse trees are obtained using the
Stanza package \citep{qi2020stanza}. 

\subsection{Semantic Textual Similarity} 

%%%%%%%%%%%%%%%%%%%%%%%%%%% 
\begin{table*}[ht] 
\centering
\caption{Spearman's ($\rho \times 100$) correlation comparison of
unsupervised methods, where the best results of each word embedding are
displayed in boldface. The number in the bracket show the performance
gain or loss of our methods as compared with WMD$_{cos}$+IDF. Results of [\textsuperscript{\textdagger}] are
taken from \citep{gao2021simcse}.}\label{table:STS_results}
{\small
\begin{tabular}{c|c|c|c|c|c|c|c|c}
\toprule
Embeddings & Methods & STS12 & STS13 & STS14 & STS15 & STS16 & STS-B & Avg. \\
\midrule
word2vec(avg.) & \multirow{6}{*}{Sent. Emb.} & 55.28 & 70.09 & 65.53 & 75.29 & 68.73 & 65.17 & 66.68\\
BERT(first-last avg.)\textsuperscript{\textdagger} &  & 39.70 & 59.38 & 49.67 & 66.03 & 66.19 & 53.87 & 55.81\\
BERT-flow\textsuperscript{\textdagger} &  & 58.40 & 67.10 & 60.85 & 75.16 & 71.22 & 68.66 & 66.90\\
BERT-whitening\textsuperscript{\textdagger} &  & 57.83 & 66.90 & 60.90 & 75.08 & 71.31 & 68.24 & 66.71\\
CT-BERT\textsuperscript{\textdagger} &  & 61.63 & 76.80 & 68.47 & 77.50 & 76.48 & 74.31 & 72.53\\
SimSCE-BERT\textsuperscript{\textdagger} &  & 68.40 & 82.41 & 74.38 & 80.91 & 78.56 & 76.85 & 76.92\\ \hline
\midrule
\multirow{7}{*}{word2vec} & WMD$_{l2}$ & 58.12 & 58.78 & 60.16 & 71.52 & 66.56 & 63.65 & 63.13\\
 & WMD$_{cos}$ & 54.82 & 61.42 & 60.71 & 72.67 & 66.90 & 62.49 & 63.30\\
 & WRD & 56.72 & 64.74 & 63.44 & 75.99 & 69.06 & 65.26 & 65.87\\
 & WMD$_{l2}$+IDF & \textbf{60.36} & 67.01 & 63.06 & 72.41 & 68.30 & 65.91 & 66.18\\ 
 & WMD$_{cos}$+IDF & 57.64 & 69.25 & 63.81 & 73.50 & 68.83 & 65.51 & 66.61\\ \cline{2-9}
 & SynWMD$_{SWF}$ & 60.24  & 74.71 & 66.10 & 75.94 & 69.54 & 66.24 & 68.80 ($\textcolor{red}{\myuparrow2.19}$)\\
 & SynWMD$_{SWF+SWD}$ & 60.30 & \textbf{75.43} & \textbf{66.22} & \textbf{75.95} & \textbf{70.06} & \textbf{66.65} & \textbf{69.10} ($\textcolor{red}{\myuparrow2.49}$)\\
\midrule
\multirow{8}{*}{BERT(first-last)} & WMD$_{l2}$ & 53.03 & 58.96 & 56.79 & 72.11 & 63.56 & 61.01 & 60.91\\
 & WMD$_{cos}$ & 55.38 & 58.51 & 56.93 & 72.81 & 64.47 & 61.80 & 61.65\\
 & WRD & 49.93 & 63.48 & 57.63 & 72.04 & 64.11 & 61.92 & 61.52\\
 & BERTScore & 61.32 & 73.00 & 66.52 & 78.47 & 73.43 & 71.77 & 70.75\\
 & WMD$_{l2}$+IDF & 61.19 & 68.67 & 63.72 & 76.87 & 70.16 & 69.56 & 68.36\\ 
 & WMD$_{cos}$+IDF & 63.79 & 69.25 & 64.51 & 77.58 & 71.70 & 70.69 & 69.59\\
 \cline{2-9}
 & SynWMD$_{SWF}$ & 66.34 & 77.08 & 68.96 & \textbf{79.13} & 74.05 & 74.06 & 73.27 ($\textcolor{red}{\myuparrow3.68}$)\\
 & SynWMD$_{SWF+SWD}$ & \textbf{66.74} & \textbf{79.38} & \textbf{69.76} & 78.77 & \textbf{75.52} & \textbf{74.81} & \textbf{74.16} ($\textcolor{red}{\myuparrow4.57}$)\\
\midrule
\multirow{8}{*}{SimCSE-BERT} & WMD$_{l2}$ & 64.66 & 79.72 & 73.12 & 81.25 & 76.69 & 77.53 & 75.50\\
 & WMD$_{cos}$ & 65.43 & 80.00 & 73.35 & 81.21 & 76.97 & 77.18 & 75.69\\
 & WRD & 64.80 & 80.97 & 74.13 & 80.71 & 76.68 & 78.47 & 75.96\\
 & BERTScore & 66.31 & 82.87 & 75.66 & 83.14 & 79.16 & \textbf{80.03} & 77.86\\
 & WMD$_{l2}$+IDF & 67.35 & 81.36 & 74.56 & 82.29 & 78.12 & 79.18 & 77.14\\ 
 & WMD$_{cos}$+IDF & 68.47 & 81.76 & 74.98 & 82.30 & 78.29 & 78.98 & 77.46\\
 \cline{2-9}
 & SynWMD$_{SWF}$ & 70.20 & 83.36 & 76.17 & 83.16 & 78.81 & 80.02 & 78.62 ($\textcolor{red}{\myuparrow1.16}$)\\
 & SynWMD$_{SWF+SWD}$ & \textbf{70.27} & \textbf{83.44} & \textbf{76.19} & \textbf{83.21} & \textbf{78.83} & 79.98 & \textbf{78.66} ($\textcolor{red}{\myuparrow1.19}$)\\
\bottomrule
\end{tabular}}
\end{table*}
%%%%%%%%%%%%%%%%%%%%%%%%%%%

\textbf{Datasets}. Semantic similarity tasks are widely used to evaluate
sentence similarity assessment methods. Here, we consider six semantic
textual similarity (STS) datasets, including STS2012-16 and
STS-Benchmark.  Sentence pairs in STS are extracted from a wide range of
domains such as news, web forum, and image captions. They are annotated
with similarity scores by humans. Each STS dataset contains several
subsets on different topics. Since it is likely to have data from
different topics in real-world scenarios, we apply the ``all setting''
evaluation for STS2012-16 as mentioned in \citep{gao2021simcse}.  The
similarity scores of the sentence pairs in different subsets are
concatenated and the overall Spearman's correlation is reported. 

\textbf{Benchmarking Methods}. We choose the following benchmarking methods.
\begin{itemize}
\item Sentence-embedding-based methods: 1) average methods: the average of
word2vec embedding \citep{mikolov2013efficient} and the average of the
first and last layers of BERT \citep{devlin2018bert}, 2) post-processing
methods: BERT-flow \citep{li2020sentence} and BERT-whitening
\citep{su2021whitening}, 3) contrastive learning methods: CT-BERT
\citep{carlsson2020semantic} and SimCSE-BERT \citep{gao2021simcse}. 
\item Word-alignment-based methods: original WMD, Word Rotator's Distance
\citep{yokoi2020word}, BERTScore \citep{bert-score}, and WMD with IDF
weights as the baselines. For exhaustive comparison, WMD using the l2
and cosine distance are both reported.  Both non-contextual and
contextual word embeddings are chosen as backbone models. They are
word2vec, pre-trained BERT, and SimCSE. 
\end{itemize}

\textbf{Experimental Setup}. In the implementation of SWF, we count word
co-occurrence in dependency parse trees if they are within 3 hops, and
set the smooth term $d=0.2$. In the implementation of SWD, we create
subtrees with child nodes of no more than 3 hops. We set $a=0.2$ for
word2vec and SimCSE word embeddings and $a=1.0$ for BERT word
embedding. 

\textbf{Isotropic Processing}. It is observed in
\cite{ethayarajh-2019-contextual} that the average cosine similarity
between randomly sampled words with pre-trained contextual word
embedding is high. This implies that pre-trained contextual word
embeddings are confined to a cone space, and they are not isotropic. It is
also shown in \cite{gao2021simcse, evalrank_2022} that the anisotropic
property of pre-trained contextual word embedding hurts its
performance in sentence similarity tasks severely. Post-processing
methods (e.g., whitening) make BERT embedding less anisotropic in the
embedding space and improves the performance in semantic similarity
tasks. Thus, when BERT embeddings are used in the experiments, we perform the whitening operation on the word level for all word-alignment-based methods. 

\textbf{Results}. We compare a wide range of methods on 6 STS datasets
and report their Spearman's correlation results in Table
\ref{table:STS_results}. For all word embeddings, WMD and WMD+IDF
perform better with the cosine distance than the l-2 distance. This
indicates that the cosine distance is a better metric for STS datasets.
Furthermore, the word flow assignment with the IDF weight can enhance
the performance of WMD. As to our proposed method, SynWMD+SWF
outperforms other alignment-based methods by a substantial margin.
SynWMD+SWF+SWD can improve SynWMD's performance even more. This is
especially obvious for word2vec and BERT embeddings.  Under the same
word embedding, SynWMD always outperforms sentence embedding methods,
including the state-of-the-art unsupervised method, SimCSE. 

%bookmark

%%%%%%%%%%%%%%%%%%%%%%%%%%%%%%%%%%%%%%%%%%%%%%%%%%%%%%%%
\begin{figure*}[tbh]
\centering % <-- added
\begin{subfigure}{0.28\textwidth}
  \includegraphics[width=\linewidth]{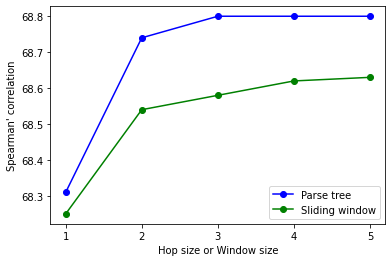}
  \caption{word2vec}
  \label{fig:word2vec}
\end{subfigure}\hfil % <-- added
\begin{subfigure}{0.28\textwidth}
  \includegraphics[width=\linewidth]{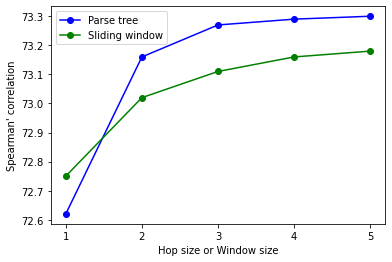}
  \caption{BERT}
  \label{fig:BERT}
\end{subfigure}\hfil % <-- added
\begin{subfigure}{0.28\textwidth}
  \includegraphics[width=\linewidth]{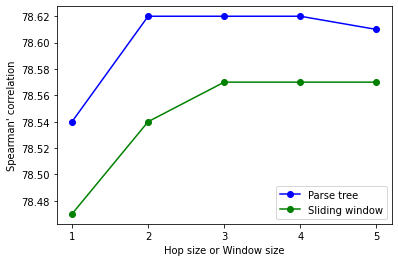}
  \caption{SimCSE-BERT}
  \label{fig:SimCSE-BERT}
\end{subfigure}\hfil % <-- added
\caption{The average Spearman's correlation curves on STS datasets as a
function of hop sizes or window sizes for three word embeddings: (a)
word2vec, (b) BERT and (c) SimCSE-BERT.}\label{fig:hopwindow-size}
\end{figure*}
%%%%%%%%%%%%%%%%%%%%%%%%%%%%%%%%%%%%%%%%%%%%%%%%%%%%%%%%

\vspace{-1ex}
\subsection{Further Analysis on STS}

We perform an ablation study on SynWMD to offer a better understanding of 
its working principle in this subsection. 

\textbf{Effect of the hop size}. We study the sensitivity of hop sizes,
$n$, in collecting word co-occurrence statistics in SWF. The blue curves
in Fig. \ref{fig:hopwindow-size} show the average performance trend with
different hop sizes on STS datasets. We see from the figure that
SynWMD+SWF with a larger hop size gives better performance. This is
because more relationships between words are incorporated for a larger
$n$. However, the performance gain saturates as $n \ge 3$.

\textbf{Difference between parse tree and linear context in SWF}. SWF
collects co-occurrence statistics from dependency parse trees, which are
well-organized structures of sentences. One can also use a sliding
window to collect co-occurrence statistics from linear contexts and
build the weighted graph. The differences between these two schemes are
shown in Fig. \ref{fig:hopwindow-size}. We see from the figure that the
dependency parse tree in SWF outperforms the sliding window. This is
because the dependency parse tree provides a powerful syntactic structure
in collecting word co-occurrence statistics. 

\textbf{Difference between subtree and n-grams in SWD.} When collecting
contextual information from words' neighbors in SWD, one can replace
subtrees with n-grams in Eq. (\ref{eq:swd}). We study the difference
between subtrees and n-grams with BERT embeddings. We generate 2-grams
and 3-grams so that the number of n-grams has the same order of
magnitude as subtrees' in our experiments. All other experimental
settings remain the same.  The performance difference between subtree
and n-grams is shown in Table \ref{table:ablation1}.  We can see from
the table that the sentence structural information does perform better 
than n-gram features. 

%%%%%%%%%%%%%%%%%%%%%%%%%%% 
\begin{table}[htb] 
\centering
\caption{Comparison of Spearman's ($\rho \times 100$) correlation of
using the subtree and the n-gram in SWD.} \label{table:ablation1}
{\small
\begin{tabular}{|c|c|c|}
\toprule
Datasets & n-gram & subtree \\
\midrule
STS12 & 66.37 & 66.64\\ 
STS13 & 78.08 & 79.40\\
STS14 & 69.36 & 69.75\\
STS15 & 79.29 & 78.82\\
STS16 & 74.41 & 75.51\\
STS-B & 74.67 & 74.93\\
Avg. & 73.70 & 74.18\\
\bottomrule
\end{tabular}}
\end{table}
%%%%%%%%%%%%%%%%%%%%%%%%%%%

\begin{figure}[t]
\centering
\includegraphics[width=0.7\linewidth]{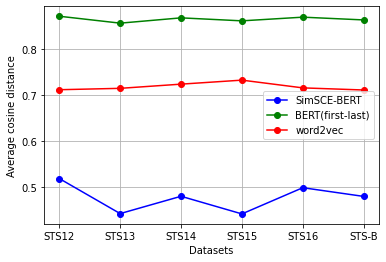}
\caption{The averaged pairwise cosine distance of words in a 
sentence of STS datasets with three embeddings.} \label{fig:ave_dist}
\end{figure}

\textbf{Effect of using different backbone word embedding models}.  As
shown in Table \ref{table:STS_results}, there is more performance
improvement by applying SWD to word2vec and BERT word embeddings but
less to SimCSE. One possible explanation for this phenomenon is that
SimCSE word embeddings in a sentence tend to be similar. When words from
a sentence have close embeddings, words and their subtrees are
expected to have close embeddings. As a result, word distances keep a
similar ratio even with the subtree distance, and results of the
constrained optimization problem, i.e., Eq. (\ref{eq:wmd}), do not
change much. To verify this point, we calculate the averaged pairwise
cosine distance of words in a sentence with three word embeddings and
show the results in Fig.  \ref{fig:ave_dist}. We see that BERT has the
largest average distance while SimCSE has the smallest. This is
consistent with their performance improvement. 

%%%%%%%%%%%%%%%%%%%%%%%%%%%%%%%%%%%%%%%
%%%%%%%%%%%%%%%%%%%%%%%%%%%%%%%%%%%%%%%

\vspace{-1ex}
\subsection{Sentence Classification}

To further validate the effectiveness of SynWMD, we perform experiments
on 4 sentence classification datasets. 

\textbf{Datasets}. We choose MR, CR, SST2, and SST5 sentence
classification datasets from SentEval \citep{conneau2018senteval}. They
are elaborated on below.
\begin{itemize}
\item MR: a movie review dataset where sentences are labeled as positive
or negative sentiment polarities. 
\item CR: a product review dataset with positive and negative
sentence reviews. 
\item SST2 \& SST5: Both are movie review datasets. SST2 has two labels
(positive and negative), while SST5 has five labels (very positive,
positive, neutral, negative, and very negative). 
\end{itemize}
Note that WMD-based methods are not suitable for the k-nearest neighbor
sentence classification with a large number of samples. For SST2 and
SST5 datasets, only test samples are used and cross-validation is
performed. They are denoted by SST2-test and SST5-test, respectively. 

\textbf{Benchmarking Methods}. We compare SynWMD with 3 other WMD-based
methods. They are: 1) original WMD, 2) Word Rotator's Distance and 3)
WMD with IDF weight. Results of WMD using the l2 and cosine distances
are reported. Word2vec is used as the backbone word embedding model in
this experiment. 

\textbf{Experimental Setup}. We set $d=0.1$ and $a=0.1$. All other
settings remain the same as those in the STS tasks. The $k$ value for
the nearest neighbor classifier is chosen from 1 to 30 to achieve the
best performance. 

\textbf{Results}: Experimental results are shown in Table
\ref{table:text_cls_results}. The cosine distance is better than the l2
distance in all four sentence classification datasets.  SynWMD
outperforms other WMD-based methods by a large margin in the k-nearest
neighbor sentence classification. 

%%%%%%%%%%%%%%%%%%%%%%%%%%% 
\begin{table}[t] 
\centering
\caption{Comparison of test accuracy for the k-nearest neighbor sentence
classification. The best results of each datasets are displayed in boldface. } \label{table:text_cls_results}
{\small
\begin{tabular}{|c|c|c|c|c|}
\toprule
Methods & MR & CR & SST2-test & SST5-test \\
\midrule
WMD$_{l2}$ & 67.68 & 73.69 & 66.12 & 31.81\\
WMD$_{cos}$ & 70.89 & 75.18 & 69.36 & 34.76\\
WRD & 73.17 & 75.74 & 72.99 & 35.25 \\
WMD$_{l2}$+IDF & 70.17 & 75.44 & 74.41 & 31.49\\
WMD$_{cos}$+IDF & 74.18 & 76.88 & 74.41 & 37.96\\
SynWMD & \textbf{76.44} & \textbf{77.08} & \textbf{77.43} & \textbf{38.28}\\
\bottomrule
\end{tabular}}
\end{table}
%%%%%%%%%%%%%%%%%%%%%%%%%%

\vspace{-3ex}
\section{Conclusion and Future Work}\label{sec:conclusion}
\vspace{-2ex}

An improved Word Mover's Distance (WMD) using the dependency parse tree,
called SynWMD, was proposed in this work. SynWMD consists of
two novel modules: syntax-aware word flow (SWD) and syntax-aware word distance (SWF). SWD examines the co-occurrence relationship between words
in parse trees and assigns lower flow to words that co-occur with other
words frequently. SWD is used to capture word importance in a sentence
using the statistics of the whole corpus. SWF computes both the distance
between individual words and their contexts collected by parse trees so
that word's contextual information and sentence's structural information
are incorporated. SynWMD achieves state-of-the-art performance in STS
tasks and outperforms other WMD-based methods in sentence classification
tasks.  As future extensions, we may extend the idea beyond
sentence-level by leveraging sentence embeddings and incorporate it in
sentence-embedding-based methods to lower the computational cost. 

\bibliographystyle{plainnat}
\renewcommand\refname{Reference}
\bibliography{mybibfile}
\end{document}